\documentclass{article}

\usepackage[final]{score_workshop}

\usepackage[utf8]{inputenc} 
\usepackage[T1]{fontenc}    
\usepackage{hyperref}       
\usepackage{url}            
\usepackage{booktabs}       
\usepackage{amsfonts}       
\usepackage{nicefrac}       
\usepackage{microtype}      
\usepackage{xcolor}         

\usepackage{multirow}
\usepackage{adjustbox}

\usepackage{amsmath,amsfonts,amssymb}

\title{Denoising Diffusion for Sampling SAT Solutions}

\author{%
  Kārlis Freivalds \\
  Institute of Electronics and Computer Science, Latvia\\
  \texttt{karlis.freivalds@edi.lv} \\
  \And
  Sergejs Kozlovičs\\
  Institute of Mathematics and Computer Science, University of Latvia\\
  \texttt{sergejs.kozlovics@lumii.lv} \\
}

\begin{document}
\maketitle

\begin{abstract}
  Generating diverse solutions to the Boolean Satisfiability Problem (SAT) is a hard computational problem with practical applications for testing and functional verification of software and hardware designs. We explore the way to generate such solutions using Denoising Diffusion coupled with a Graph Neural Network to implement the denoising function. We find that the obtained accuracy is similar to the currently best purely neural method and the produced SAT solutions are highly diverse, even if the system is trained with non-random solutions from a standard solver.  
  
\end{abstract}

\section{Introduction}
Boolean Satisfiability Problem (SAT) is a significant NP-complete problem with numerous practical applications -- product configuration, hardware verification, and software package management, to name a few. There is no known single efficient algorithm that solves every SAT problem, but heuristics have been developed that are well-suited to instances arising in practice. Here we consider an even harder task of sampling solutions of a SAT instance which is the key component for functional verification of software and hardware designs. Most SAT solvers return the same solution if run repeatedly; therefore, special approaches are needed to obtain several solutions, especially if it is required to sample more or less uniformly from the solution space. UniGen \citep{CMV14, SGM20} and QuickSampler \citep{dutra2018efficient} are some algorithms for this purpose. UniGen guarantees approximate uniformity of the sampled solutions but is very slow in certain cases. So, we are interested in whether the application of neural networks and denoising diffusion can produce diverse samples in less time. 

Deep learning methods can accelerate solving NP-hard problems. Purely neural SAT solvers have been developed \citep{SelsamLBLMD19, amizadeh2019pdp, ozolins2021goal} that can solve small instances. Hybrid solvers that provide neural heuristics within existing solvers \citep{selsam2019guiding, han2020enhancing} are more practical for real-life instances. Also,  deep learning methods are used for Combinatorial Optimization \citep{NEURIPS2018_8d3bba74, bengio2021machine, cappart2021combinatorial} including MIP \citep{nair2020solving}, TSP \citep{gaile2022unsupervised} and VRP \citep{NEURIPS2018_9fb4651c}.  

Denoising diffusion \citep{yang2022diffusion}, coupled with neural networks, provide a learnable way for sampling from a given distribution. Diffusion models have achieved state-of-the-art results for image generation \citep{ho2022cascaded, vahdat2021score, rombach2021high}. Diffusion has been applied to discrete binary \citep{sohl2015deep} and categorical \citep{hoogeboom2021argmax} data, text generation \citep{austin2021structured}, symbolic music generation \citep{mittal2021symbolic}, graph generation \citep{niu2020permutation}, and autoregressive modelling \citep{hoogeboom2021autoregressive}. But we are not aware of any attempt to use it for sampling SAT solutions. 

In this paper, we apply denoising diffusion to sample solutions of SAT. We use Categorical Diffusion introduced by \citep{hoogeboom2021argmax} and couple it with QuerySat \citep{ozolins2021goal} graph neural network to implement the denoising function. We find that this approach achieves solving accuracy comparable to the state-of-the-art purely neural SAT solver (QuerySat) and finds diverse solutions of similar characteristics as UniGen up to 8000 times faster, in some cases.


\section{Background: Diffusion Models}
\label{sec:background_diff}
Given data $x_0$(variable assignment in our case), a diffusion model \citep{sohl2015deep} consists of predefined variational distributions $q(x_t|x_{t-1})$ that gradually introduce noise over time steps $t \in \{1,...,T\}$. The diffusion trajectory is defined such that $q(x_t|x_{t-1})$ adds a small amount of noise around $x_{t-1}$.
This way, information is gradually destroyed and at the final time step, $x_T$ carries almost no information about $x_0$. A nice property of the diffusion process is that it can be reversed if the gradient of the distribution can be estimated which is often expressed as a function that denoises the data. Usually, the Normal distribution for noise is used which is simple to work with and produces excellent results for images \citep{ho2022cascaded} and sound \citep{Diffwave}.

To deal with the discrete SAT solution values, here we employ diffusion for categorical data, namely the Multinomial Diffusion \citep{hoogeboom2021argmax}. Having $K$ categories (2 in our case -- True and False), $x_t$ is encoded as one-hot vector for each variable. The multinomial diffusion process is defined using a categorical distribution that has a small probability $\beta_t$ of resampling a category uniformly and a large $(1-\beta_t)$ probability of sampling the previous value $x_{t-1}$:

\begin{equation}
q(x_t|x_{t-1}) = \mathcal{C}(x_t|(1-\beta_t)x_{t-1}+\beta_t/K),
\end{equation}

where $\mathcal{C}$ denotes a categorical distribution with probability parameters after $|$. For such diffusion process the probability of any $x_t$ given $x_0$ is expressed as: 
\begin{equation}
q(x_t|x_0) = \mathcal{C}(x_t|\bar{\alpha}_t x_0+(1-\bar{\alpha}_t)/K),
\label{eq:qsample}
\end{equation}
where $\alpha_t=1-\beta_t$ and $\bar{\alpha}_t=\prod_{\tau=1}^t{\alpha_\tau}$. For the reverse distribution step, we follow the common practice to parametrize it using $x_0$.  According to \citep{hoogeboom2021argmax}, the distribution for the previous time step $t-1$ can be computed from the value $x_t$ at the next step and the initial value $x_0$ as:
\begin{equation}
\begin{split}
q(x_{t-1}|x_t, x_0) = \mathcal{C}(x_{t-1}|\theta_{post}(x_t, x_0)),
\text{where } \theta_{post}(x_t, x_0)=\tilde{\theta}/\sum_{k=1}^K{\tilde{\theta}_k},\\
\text{and } \tilde{\theta}=[\alpha_t x_t + (1-\alpha_t)/K]\odot[\bar{\alpha}_{t-1} x_0 + (1-\bar{\alpha}_{t-1})/K]
\end{split}
\label{eq:3}
\end{equation}

During the reverse process, an approximation $\hat{x}_0$ is used instead of $x_0$ which is produced by a neural network $\mu$: $\hat{x}_0=\mu(x_t, \bar{\alpha}_t)$. The neural network is trained by feeding it with the cumulative noise magnitude\footnote{\citet{hoogeboom2021argmax} parametrize the neural network with $t$, instead. This is equivalent once we fix the noise schedule.} $\bar{\alpha}_t$ and noisy sample $x_t$ which is produced by the forward diffusion and asking the network to produce a clean sample $\hat{x}_0$.The loss function for training is the KL divergence between the true distribution and the predicted one: 
\begin{equation}
KL(\mathcal{C}(\theta_{post}(x_t, x_0))|\mathcal{C}(\theta_{post}(x_t, \hat{x}_0)))
\label{eq:KL}
\end{equation}

\section{Background: GNNs for Solving SAT}
\label{sec:background_sat}
We consider SAT formulas in conjunctive normal form and represent them as a bipartite variables-clauses graph, likewise known as an SAT factor graph \cite{biere2009handbook}. In such a graph, the edge between variables and clauses graph exists whenever the variable is present in the clause. Two types of edges are used to distinguish a variable from its negation.
Variables-clauses graph of SAT formula with $n$ variables and $m$ clauses can be represented as a sparse $n \times m$ adjacency matrix. 

Graph Neural Networks(GNN), especially message-passing networks, are commonly used for solving graphs-based tasks. Such network architectures are simple, yet have strong inductive bias over the graph \cite{zhou2018graph}. For boolean satisfiability solving, the currently best neural architecture is QuerySat \citep{ozolins2021goal} which is capable of predicting solutions to random 3-SAT formulas up to 400 variables. QuerySat is a recurrent GNN, specially tailored for SAT solving, which at each iteration exchanges the information between variables and clauses, gives a candidate prediction, and evaluates how far it is from a valid solution. It is trained with 32 recurrent iterations but at inference time the number of iterations can be increased to yield better results. The prediction at the final iteration is used as the result.  
We use QuerySat architecture with 32 recurrent iterations as the  $\mu$ network for denoising.

\section{Diffusion for SAT Sampling}\label{sec:satsolving}
To train the neural network $\mu$, we generate a dataset with SAT formulas and their solutions. The solution is one-hot encoded and treated as $x_0$ but the SAT instance is used to build the GNN network so the $\mu$ function gets conditioned on the given instance. Then for each example we randomly sample  $\bar{\alpha}_t$ according to noising schedule $\bar{\alpha}_t = 1-\sqrt{t/T}$ and obtain $x_t$ by sampling $q(x_t|x_0)$ as given in Eq.~\ref{eq:qsample}. The neural network $\mu(x_t, \bar{\alpha}_t)$ is trained using KL divergence given by Eq.~\ref{eq:KL} to estimate $x_0$. 

After the model has been trained, we can use it for sampling a solution to the given SAT formula. We employ the standard denoising diffusion algorithm. Given the total number of steps $T$, the algorithm works backward from step $T$ toward step 1. The distribution $x_T$ is created with equal probabilities of True and False for each variable and then at each step $t$ a sample from it is drawn which is presented to the neural network $\mu$. The neural network returns the estimated probabilities of SAT variable assignment $\hat x_0$. The value of the previous timestep $x_{t-1}$ is obtained by sampling the distribution $q(x_{t-1}|x_t, \hat{x}_0)$ given by Eq.~\ref{eq:3}. 
The algorithm returns the assignment $\hat x_0$ with argmax applied to each variable. 


\section{Evaluation}
\label{sec:results}

We evaluate the proposed SAT sampling algorithm on two datasets: random 3-SAT and 3-Clique instances. The 3-SAT instances consist of random clauses of 3 literals at the satisfiability threshold, i.e., with the clause/variable ratio of about 4.3. We also test on 3-SAT instances with clause/variable ratio 3. The 3-Clique instances represent SAT encoding of the task to find triangles in a graph. See Appendix \ref{app:data} for the dataset details. 



At first, let us evaluate the accuracy of the produced solutions i.e. the fraction of the instances that are fully solved. For accuracy evaluation, we consider an instance solved if a solution is found at any of the diffusion steps, this increases our chances to spot a solution since a great amount of randomness is involved in the diffusion process. We compare our solution, named DiffusionSat, with QuerySat. For DiffusionSat we consider two choices for the number of diffusion steps -- 32 or 128. For QuerySat we use two setups of 32 or 4K recurrent iterations, as it was done in \citep{ozolins2021goal}. Note that the amount of computing for QuerySat 4K is roughly the same as for DiffusionSat 128. Since the small instances used for training are solved almost perfectly, for accuracy evaluation, we use about 4x larger ones. 

We consider two solvers for generating the solutions that are used for DiffusionSat training -- Glucose4 \citep{audemard2018glucose} and UniGen3 \citep{SGM20}. Glucose is a standard SAT solver that does not guarantee any particular distribution of the produced solutions but UniGen provides approximately uniformly distributed solutions. Diffusion models are intended to reproduce the same distribution of samples that was supplied for training, so we may get different results in both cases.

The accuracy results are given in Table~\ref{accuracy-table}. It depicts the average and the standard deviation of 5 test runs. We see that QuerySat is better for 3-SAT but DiffusionSat is better for 3-Clique. Notably, DiffusionSat 128 achieves 100\% solved instances of 3-Clique where QuerySat 4K gets about 95\% while both these cases use the same amount of computing. Also, we can notice that increasing the number of diffusion steps increases the accuracy. Notice that UniGen always produces correct solutions, so it is not evaluated.


\begin{table}
  \caption{Test accuracy (higher is better) as percent of fully solved instances.}
  \label{accuracy-table}
  \centering
  \begin{tabular}{lccc}
    \toprule
    \multirow{2}{*}{Method} & \multicolumn{2}{c}{3-SAT} & \multicolumn{1}{c}{3-Clique}                   \\
    \cmidrule(lr){2-3}\cmidrule(lr){4-4} 
         & cl/var $\approx$ 4.3     & cl/var = 3.0 \\
    \midrule
    QuerySat (32 iterations) & $61.9 \pm 5.2$  & $\textbf{100} \pm 0$  & $82.0 \pm 4.7$   \\
    QuerySat (4K iterations) & $\textbf{93.3} \pm 3.2$  & $\textbf{100} \pm 0$  & $94.7 \pm 4.6$   \\    
    DiffusionSat Glucose (32 steps) & $45.3 \pm 0.9$ & $95.2 \pm 0.4$ & $67.5 \pm 2.5$      \\
    DiffusionSat UniGen (32 steps) & $40.8 \pm 0.9$ & $99.9 \pm 0.0$ & $99.1 \pm 0.1$  \\
    DiffusionSat UniGen (128 steps) & $47.1 \pm 0.7$ & $\textbf{100} \pm 0$ & $\textbf{100} \pm 0$  \\    
    \bottomrule
  \end{tabular}
\end{table}

In order to evaluate sample diversity, we produce 100 solutions of each instance and count how many of them are unique. Incorrect solutions, which may appear due to non-perfect accuracy of DiffusionSat, are filtered out. The evaluation is performed on 3-SAT instances of size 5--100 variables and on 3-Clique instances obtained from graphs with 4 to 40 vertices. Our first evaluation revealed that 3-SAT at the satisfiability threshold and 3-Clique instances have only a few solutions -- they have been designed to be maximally hard and are not well-suited to evaluate diversity. So, we included a dataset with 3-SAT instances with clause/variable ratio of 3 which have many solutions. 

Table~\ref{unique-table} shows the solution diversity. We see that QuerySat produces only a few different solutions, although it employs some randomness  internally. The best results are produced by DiffusionSat, even surpassing UniGen. Interestingly, there is not much difference whether DiffusionSat has been trained with diverse solutions or with those coming from a standard SAT solver -- both produce slightly better diversity than UniGen. 

\begin{table}
  \caption{Percentage of unique solutions (higher is better) from 100 samples. The lowest possible value is 1 that occurs when only one unique solution is found in 100 runs.}
  \label{unique-table}
  \centering
  \begin{tabular}{lccc}
    \toprule
    \multirow{2}{*}{Method} & \multicolumn{2}{c}{3-SAT} & \multicolumn{1}{c}{3-Clique}                   \\
    \cmidrule(lr){2-3}\cmidrule(lr){4-4} 
         & cl/var $\approx$ 4.3     & cl/var = 3.0 \\
    \midrule
    QuerySat (32 iterations) & $6.7 \pm 1.1$  & $6.3 \pm 0.7$  & $1.2 \pm 0.1$   \\
    DiffusionSat Glucose (32 steps) & $56.6 \pm 7.1$ & $95.0 \pm 3.1$ & $\textbf{5.9} \pm 0.2$      \\
    DiffusionSat UniGen (32 steps) & $\textbf{61.5} \pm 7.6$ & $\textbf{95.9} \pm 2.8$ & $2.6 \pm 0.2$  \\
    UniGen & $49.4 \pm 3.0$ & $92.6 \pm 5.9$ & $1.4 \pm 0.1$  \\
    \bottomrule
  \end{tabular}
\end{table}

To get more insight how diverse are the solutions, we inspect how many variables are equal in two solution samples on 3-SAT instances, see Appendix~\ref{app:results}. For the clause/variable ratio of 4.3 about 85\% variables are equal and for ratio 3 -- about 65\%. Notice that completely random assignment would yield 50\% but that may not be a valid solution. We find that DiffusionSat and UniGen perform pretty similarly regarding this metric.  

We evaluated the time taken to produce one solution sample on 3-SAT instance with 100 variables. DiffusionSat (with 32 diffusion steps) processes one batch of about 95 instances in 5.7 seconds. That means, one instance takes 0.06s. UniGen takes about 0.22s per instance of clause ratio 4.3 but much longer -- about 500s (with high variance from 70s to 1300s) for clause ratio 3, see also Appendix~\ref{app:results}. When compared, DiffusionSat produces a solution more than 8000 times faster for the latter case (although using GPU and batched instances).

\section{Conclusion}
\label{sec:conclusions}
We have shown how to combine denoising diffusion with GNN and apply it for generating SAT solutions. From theoretical point of view, if trained with uniformly distributed samples and the neural network being able to perfectly learn the denoising function, the samples produced by diffusion are also uniformly distributed. Both these requirements are hard to satisfy in practice, but we have shown that DiffusionSat generates highly diverse samples of similar characteristics than UniGen which is known to generate nearly uniform distribution. Moreover, there is no need to use uniform distribution for training, as training on solutions from a standard SAT solver gives equally good results. Besides, for 3-SAT formulas with clause/variable ratio 3, DiffusionSat is much faster that UniGen.

There are some limitations of this work: (a) We have tested only two properties of diversity, more thorough tests should be done to draw conclusions about uniformity. Also, more datasets should be considered. (b) Training sets with larger instances may be infeasible to generate, QuerySat has an advantage here, not requiring to know the solutions for training. (c) DiffusionSat is an incomplete solver -- it cannot prove that an instance is UNSAT. (d) SAT instances up only to 50K clauses can be solved with the current A6000 GPU(with 48GB memory).

\begin{ack}
This research is supported by the Latvian Council of Science, project lzp-2021/1-0479. Computing resources are supported by NVIDIA Academic Grant "Deep Learning of Algorithms". 
\end{ack}

\bibliography{bibliography.bib}

\clearpage
\appendix

\section{Appendix: Datasets}
\label{app:data}
3-Clique and 3-SAT tasks are generated using the CNFGen library \cite{lauria2017cnfgen}, which allows encoding several popular problems as SAT instances. We generate hard 3-SAT instances at the satisfiability boundary where the relationship between the number of clauses ($m$) and variables ($n$) is $m = 4.258n + 58.26n^{-\frac{2}{3}}$ \cite{crawford1996experimental}. The train and validation sets of 3-SAT tasks consist of formulas with 5 to 100 variables. Since the small instances used for training are solved almost perfectly, for accuracy evaluation, we use larger ones -- up to 405 variables. Such larger test instances were also used in the QuerySat paper \citep{ozolins2021goal} and show the generalization ability of the methods. The test set for sample diversity evaluation has instances of the same size as used for training, i.e. with 5 to 100 variables. For diversity evaluation, we have another test set with clause-to-variable ratio $m = 3n$. These instances are much easier and have many solutions. 

For the 3-Clique task, we generate Erdős–Rényi graphs with edge probability $p$. The goal of this task is to find all triangles in the graph. We use $p=3 ^ \frac{1}{3} / (v(2 - 3v + v^2))^ {\frac{1}{3}}$, where $v$ is the vertex count in the graph. Such generation produces graphs in which a few triangles are expected. The graphs are then encoded as SAT instances using the CNFGen library. Train and validation sets consist of SAT encodings of graphs with 4 to 40 vertices. They end up encoded into SAT formulas of sizes up to 120 variables and 7K clauses. The test set for accuracy evaluation has graphs with 4 to 100 vertices. The test set for diversity evaluation has graphs with 4 to 40 vertices.

\clearpage
\section{Appendix: Implementation and Training}
\label{app:training}
We have implemented DiffusionSat in Tensorflow based on the official code of QuerySat\footnote{\href{https://github.com/LUMII-Syslab/QuerySAT}{https://github.com/LUMII-Syslab/QuerySAT}}. From there, we employ the GNN neural backbone to implement the $\mu$ network, the dataset generation and accuracy evaluation code. So, the accuracy results of DiffusionSat and QuerySat are obtained in the same conditions and are directly comparable. The backbone network that is employed inside DiffusionSat has 32 recurrent iterations that is the same as used for training QuerySat. 

For generating SAT solutions we use Glucose4 solver from python-sat\footnote{\href{https://pypi.org/project/python-sat/}{https://pypi.org/project/python-sat/}} library and UniGen3\footnote{\href{https://github.com/meelgroup/unigen}{https://github.com/meelgroup/unigen}}. 
Only satisfiable instances are used in training. 

Training is performed for 166K steps with AdaBelief optimizer \citep{zhuang2020adabelief} having learning rate 3e-3. Several instances are batched together to increase training and inference speed. Training on a single Nvidia RTX A6000 GPU takes about 1 day. 

To evaluate 3SAT instances with clause/variable ratio 3, we do not perform separate training, we use the same model trained for the  4.3 ratio -- it works very well.

\clearpage
\section{Appendix: Additional Results}
\label{app:results}

In order to evaluate diversity, we inspected how many variables are equal in two solution samples on 3-SAT instances. The results for DiffusionSat and UniGen are quite similar.

\begin{table}[h]
  \caption{Percentage how many variables are equal between two solution samples (lower is better; random variable assignments would yield 50\%). }
  \label{var-table}
  \centering
  \begin{tabular}{lcc}
    \toprule
    \multirow{2}{*}{Method} & \multicolumn{2}{c}{3-SAT}                    \\
    \cmidrule(lr){2-3} 
         & cl/var $\approx$ 4.3     & cl/var = 3.0 \\
    \midrule
    DiffusionSat Glucose (32 steps) & $86.0 \pm 1.7$ & $62.4 \pm 6.6$      \\
    DiffusionSat UniGen (32 steps) & $86.4 \pm 2.0$ & $65.0 \pm 5.8$   \\
    UniGen & $83.1 \pm 16.4$ & $63.2 \pm 7.4$  \\
    \bottomrule
  \end{tabular}
\end{table}

\begin{figure}[h]
    \centering
    \includegraphics[width=0.7\textwidth]{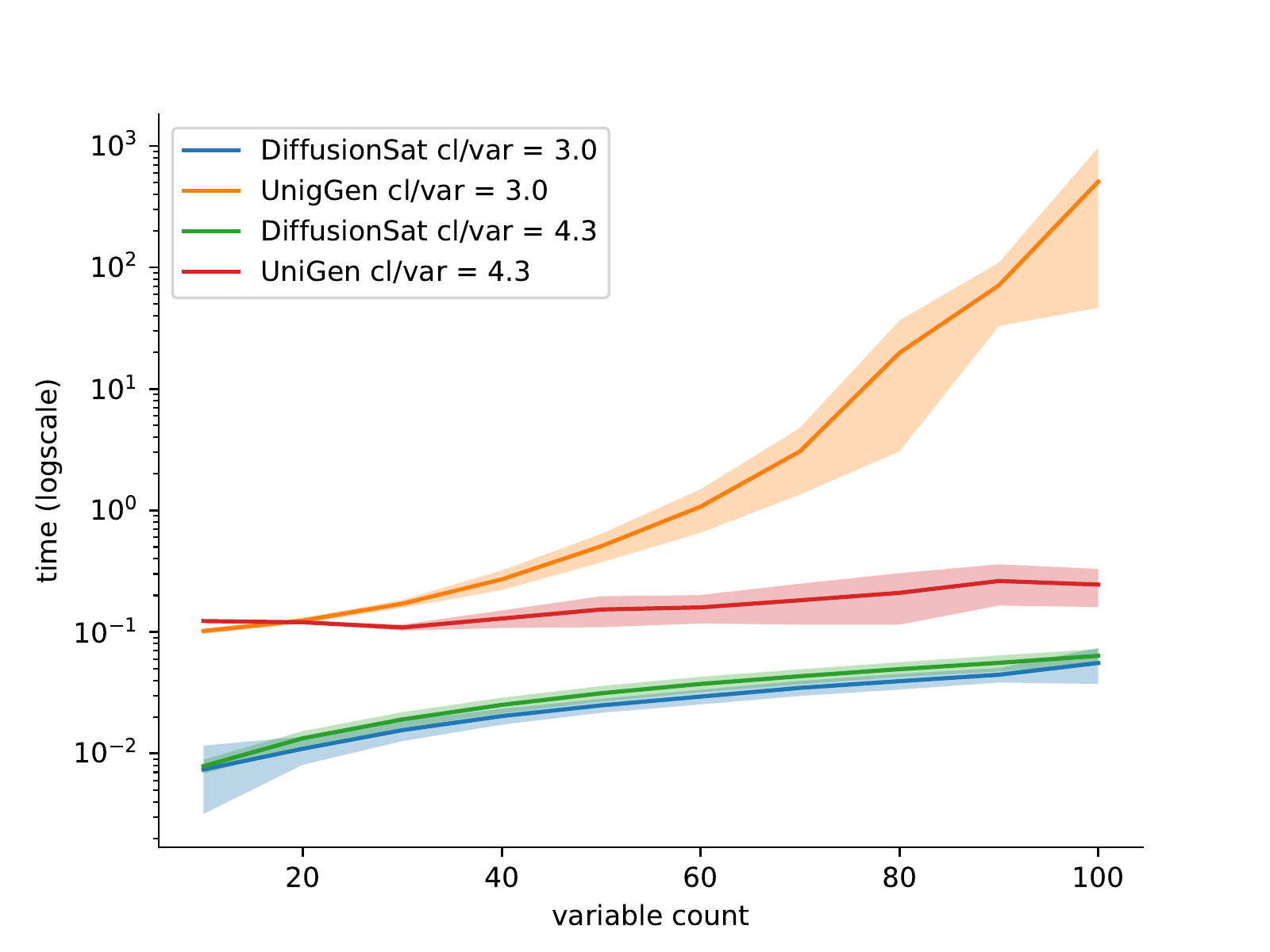}
    \caption{Sampling time depending on the formula size. Note that time is presented in log-scale.}
    \label{fig:time}
\end{figure}

The running time comparison between DiffusionSat and UniGen is given in Fig~\ref{fig:time}. This graph shows the time to produce one solution sample of a 3-SAT instance with increasing variable count.  DiffusionSat is run on a A6000 GPU, 32 diffusion steps are used, instances are batched and time is reported by dividing the wall-clock time with the instance count in the batch. UniGen is run on a CPU, one instance at a time. In such setup, we can see that DiffusionSat is much faster and scales linearly depending on the instance size. UniGen, in contrary, shows at least exponential scaling on instances of clause/variable ratio 3 reaching already about 500 seconds on a 100 variable instance.

\end{document}